# Plotting Markson's "Mistress"

**Conor Kelleher**
School of Computer Science
University College Dublin
Dublin, Ireland
conor.kelleher.1@ucdconnect.ie

**Mark T. Keane**
Insight Data Analytics Centre
University College Dublin
Dublin, Ireland
mark.keane@ucd.ie

## Abstract

The post-modern novel "Wittgenstein's Mistress" by David Markson (1988) presents the reader with a very challenging non-linear narrative, that itself appears to one of the novel's themes. We present a distant reading of this work designed to complement a close reading of it by David Foster Wallace (1990). Using a combination of text analysis, entity recognition and networks, we plot repetitive structures in the novel's narrative relating them to its critical analysis.

## 1 Introduction

*Certain novels not only cry out for critical interpretations but actually try to direct them.*
(Wallace, 1990, p. 244)

In recent years, a considerable body of research has contrasted the close reading of traditional literary criticism with distant readings based on text and network analytics (e.g., Jänicke et al., 2015; Moretti, 2005, 2013; Serlen, 2010). In this paper, we take the ecumenical position that distant readings can surface unappreciated patterns in a literary work to support close readings (c.f., Coles et al., 2014; Jänicke et al., 2015). For instance, a number of distant readings have supported close readings by providing abstractions showing the social networks of characters involved in a literary work (Agarwal et al., 2012; Grayson et al., 2016; Elson et al., 2010). The present work has, in part, been inspired by these distant readings, thought it explores phrase repetition and association networks. Specifically, we analyse the structure of repetitions in the post-modern novel "Wittgenstein's Mistress" (1988) by the American author, David Markson to support a close reading of the novel by David Foster Wallace (1990).

### 1.1 Close Reading "Wittgenstein's Mistress"

David Markson's (1988) novel "Wittgenstein's Mistress" (WM) is renowned for it use of an experimental, non-linear narrative structure. The novel deals with the thoughts of a woman – simply known as Kate – who appears to be the last living person on earth. The narrative captures the streamed consciousness of her thoughts about art, her travels though her empty world, and her reflections on the imprecision of language. The novel has no chapters but consists of almost 4,000 short paragraphs many of which are single sentences, exemplified by following excerpt:

*Gaetano Donizetti being still another person whom I otherwise might have mixed up with Vincenzo Bellini. Or with Gentile Bellini, who was also Andrea Mantegna's brother-in-law, being Giovanni Bellini's brother.*

*Well I did mix him up. With Luigi Cherubini.*

*Music is not my trade.*

*Although Maria Callas singing in that particular scene has always sent shivers up and down my spine.*

*When Van Gogh was mad, he actually once tried to eat his pigments.*

*Well, and Maupassant, eating something much more dreadful than that, poor soul.*

*The list becomes distressingly longer.*

(Markson, 1988, p. 111)

One major feature of the narrative is Kate's repetitive turn of thought. Regularly, phrases and paragraphs are repeated throughout the novel, or the same entities, people and places are mentioned over and over again. To the reader the narrative often seems like a long piece of classical music where various motifs and themes are repeated and returned to at definite intervals; but, this structure is almost impossible to grasp because the "musical phrases" are "word phrases", repetitions of named people (e.g., Maupassant), places (e.g., Rome) and turns-of-phrase (e.g., 'so to speak'). Understanding this repetitive stream of consciousness and its significance is perhaps one of the main critical challenges of the novel. In his critical commentary, David Foster Wallace (1990) says:

> *You could call this technique 'deep nonsense', meaning I guess a linguistic flow of strings, strands, loops and quiffs that through the very manner of its formal construction flouts the ordinary cingula of 'sense' and through its defiance of sense's limits manages somehow to 'show' what cannot ordinarily be 'expressed'.*
> (Wallace, 1990, p. 248)

One way to think about the current distant reading is that it is an attempt to surface the "formal construction" of WM, to better understand its message.

A second major feature of the novel is it concern with language itself, or rather many of the concerns raised by Wittgenstein in his early and late philosophy of language. Wallace argues that the book could be viewed as "a weird cerebral *roman a clef*" that attempts to dramatise "the very bleak mathematical world (of) Wittgenstein's *Tractus*"; that it tries to show what it would like "if somebody really had to live in a *Tractatus*ized world" (Wallace p. 246). In particular, Wallace points to the atomistic, independence of words and what they refer to, in this world, quoting Wittgenstein:

> *The world is everything that is the case; the world is nothing but a huge mass of data, of logically discrete facts that have no* intrinsic *connection to one another. Cf the* Tractus *1.2: 'The world falls apart into facts…' 1.2.1" 'Any one fact can be either the case, or not the case, and everything else remains the same'.*
> (Wallace, 1990, p. 254)

Faced with this world where there is no connection between 'facts', Wallace points out "that Kate's textual obsession is simply to find connections between things, any strands to bind the historical facts & empirical data that are all her world comprises" (pp. 254-255).

It is clear that a close reading of WM presents real challenges for anyone attempting to understand its structure. Hence, in this paper, have attempted a distant reading using text analytic techniques to surface the novel's repetition-structures to support Wallace's critical analysis.

## 2 Analysing Markson's Mistress

We believed that the key to the narrative structure of *Wittgenstein's Mistress* lies in surfacing the repetitive aspects of the narrative, taking the novel's short paragraphs as a sort of "fundamental unit of analysis". Hence, we progressively analysed the text in four distinct ways to surface these repeating structures, exploring:

- *Repeated Phrases*: generating all n-grams of differential lengths from (2-32) we identified repeating unique, maximal strings of characters in the text (aka repeated phrases)
- *Named Entities*: taking these maximal n-gram strings, we extracted all the entities mentioned in them, as the narrative often relies heavily on repeated references to people, places and things
- *Networks of Association*: we then used these named entities to build associative networks, on basis of how "close" they were to one another in these repeating text-fragments (see our definitions of "close" in later section)
- *Sequences of Repetitions*: finally, we examined whether these networks (clumps of associated entities) occurred in definite sequences; that is, was there a definite tendency for one cluster to follow another in patterned sequences, to determine whether we could sketch the "DNA of the novel"

In the following subsections, we elaborate the specific methods used to perform these different analyses and what we discovered using them.

### 2.1 Finding Repeated Phrases

Intuitively, reading WM one has a strong sense that certain phases are consistently re-used throughout the text. However, is very hard to get a sense of the scale of this repetition or, indeed, its structure without resorting to text analytic methods. So, to capture the extent of the repetition we performed an analysis of the novel's repeated phrases.

We transformed a PDF of the novel into a plain text file retaining its punctuation and paragraph boundaries. The novel has 3,804 paragraphs, 4,352 sentences and a total of 81,970 words, referring to approximately 3,058 named entities (*people, places, organisations, etc.*) of which 462 are unique, named entities. Treating this text as a long string of words, we used different-sized, n-gram windows and slid them over the text to find repeated phrases (using the NLTK tokenisation packages and our own n-gram program). Using this method all possible n-grams were computed from 2-grams up to 32-grams halting the search when no further repetition was found in the n-gram set. As we looped through n-grams of different window sizes, we retained repeated n-grams and discarded any that only occurred once. Furthermore, across n-gram sets of different sizes we discarded n-grams that had word-sequences that were subsets of longer n-grams, so we ended up with a list of all the unique, maximally-sized n-grams that had been repeated at least two times or more in the novel.

| N-gram | N |
|---|---|
| Was it really some other person I was so anxious to discover when I did all of that looking or was it only my own solitude that I could not abide ? | 32 |
| Although doubtless when I say they are half empty I should really be saying they are half filled since presumably they were totally empty before somebody half filled them | 29 |
| Still I find it extraordinary that young men died there in a war that long ago and then died in the same place three thousand years after that | 28 |
| Even if a part I have always liked is when Orestes finally comes back after so many years and Electra does not recognize her own brother | 26 |
| Then again it is not impossible that they were once filled completely becoming half empty only when somebody removed half of the books to the basement | 25 |

Table 1: Top-5 Longest Repeated N-grams

In our n-gram search upto 32-grams, we identified a large number of unique, maximal n-grams; there were 4,503 phrases, different-sized n-grams that were repeated at least twice in the text (see Tables 1 and 2).

Table 1 shows the top-5 n-grams by length found in the text (i.e., number of words they contain), each of which only occurred twice in the novel. Table 2 shows the top-10 most frequently repeating n-grams found in the text. Notably, the most frequent maximal n-grams are typically quite short.

This analysis gives us a first sense of the extent of the repetition in the text, indicating that this repetition is being deployed systematically to achieve some narrative effect. Indeed, of the 3,804 paragraphs in the novel, 3,323 contain at least one or more maximal n-gram with >3 words. In the next sub-section, we get a sense of the systematicity that may lie in this repetition by identifying the named entities involved in these repeated sequences.

| N-gram | N |
|---|---|
| now that I think about it | 8 |
| There would appear to be no | 7 |
| doubtless I would not have | 7 |
| Even if I have no | 7 |
| When one comes right down to it | 6 |
| But be that as it may | 6 |
| As a matter of fact the | 6 |
| As a matter of fact what | 6 |
| And to tell the truth I | 6 |
| God the things men used to do | 5 |

Table 2: Top-10 Frequently Repeated N-grams

### 2.2 Identifying Named Entities

To further analyse the repeated text fragments found in the n-gram analysis, we used the NLTK's Stanford Named Entity Recogniser (NER) package to find all the named entities mentioned in the set of maximal n-grams; though we needed to manually extend its functionality to remove incorrect identifications and to add missing entities (to deal with problems like "William de Kooning" and "de Kooning" being treated as separate entities). The full text yielded 462 unique entities (from 3,804 paragraphs) and when we searched our maximal n-gram set, we found that 117 of these entities were mentioned in the paragraphs containing repeated phrases (341 paragraphs). We used this set of entities as a basis for examining the repetitions of people, places and things in networks of association.

### 2.3 Networks of Association

To get a real sense of the structure of the repetition we constructed networks of association between the 314 unique named entities that were mentioned in our n-gram text-fragments. We build two different networks based on different measures of association (for other possibilities see Grayson et al., 2016). First, we build networks for entities that were *associated by virtue of being mentioned in the same paragaph* (see Figures 1-5). Second, we build networks for entities that were *associated by virtue of being mentioned in the same and/or the immediately following paragraph* (see Figure 6). These networks revealed that there are definite and distinct repetitive-structures of association in the novel.

For example, Figure 2 shows one of the smaller networks – which we call the Homeric Network --- showing associations between a variety of Greek-related entities, reflecting references to Homer (e.g., Illiad, Sirens, Odysseus). Note, that the thickness of the link in this network indicates the frequency with which these two entities were repeatedly associated within the paragraphs of the novel. So, for instance, the link between Achilles and Odysseus is thicker because they are mentioned together in paragraphs several times.

Performing this type of analysis we found five disjoint networks of different sizes between other named entities that we have titled after their dominant associations: (i) Paris Network (Figure 1), (ii) Homeric Network (Figure 2), the Rome Network (Figure 3), (ii) the Spanish Network (Figure 4), (iii) the Gallery Network (Figure 5).

These five networks show Kate's swirling, repetitive thoughts about these entities when we define the association very tightly (i.e., same paragraph mentions). They show a considerable amount of local coherence. However, if we widen the definition of "association" to one-paragraph further out, we find that all of these networks become inter-connected. Figure 6 shows the swirling complexity of the associations between all of the named entities as the more local networks get connected together. Note, again, that the width of the links in this network indicates repeated associations.

This wider definition of association shows that that the original in-paragraph networks are reinforced and expanded with new entities; though they still retain their identity as distinct networks. Figure 6 also shows that there are hub-like entities (e.g., Dylan Thomas, Nightwatch, La Mancha) that play a role in cross-connecting the in-paragraph networks; perhaps indicating that these entities are pivotal in Kate's thinking (n.b., Markson claimed Dylan Thomas as a major influence). Such hubs could be further elucidated by a pagerank analysis.

## 3 What Does it All Mean?

We began this paper by proposing that Wallace's close reading of WM could be complemented by a distant reading of the novel. We have shown that a text analytic and network analysis of the repeated mentions of people, places and things in the novel can reveal some definite structures. But, what does it all mean?

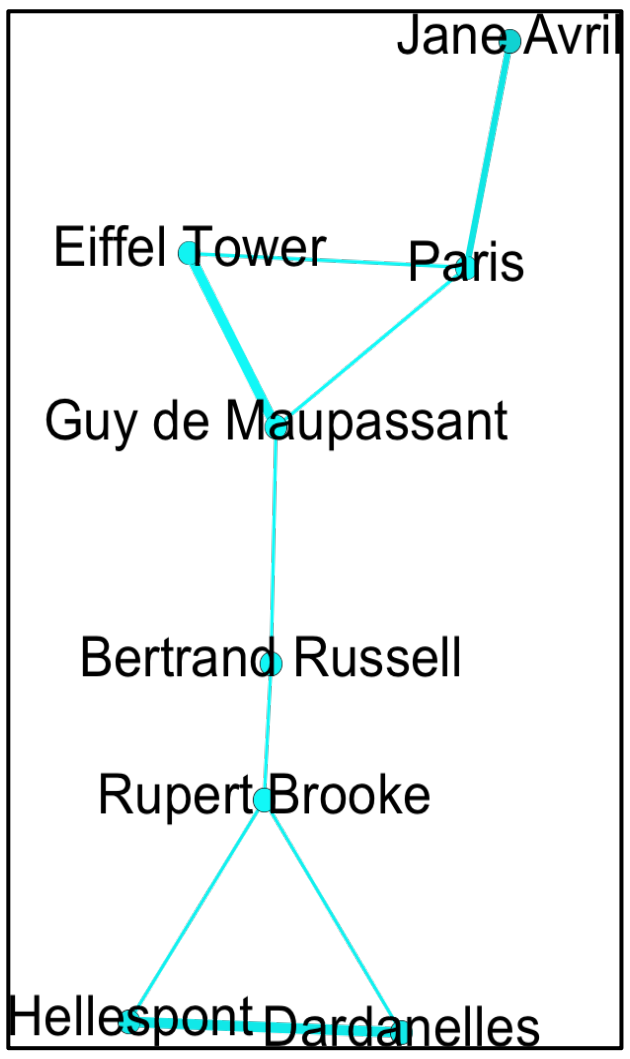

Figure 1: The Paris Network

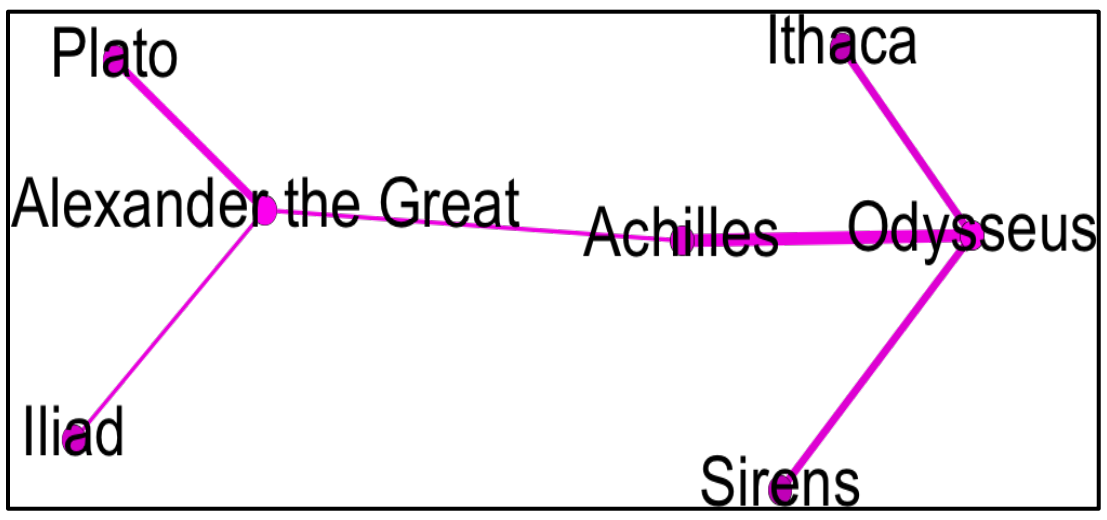

Figure 2: The Homeric Network

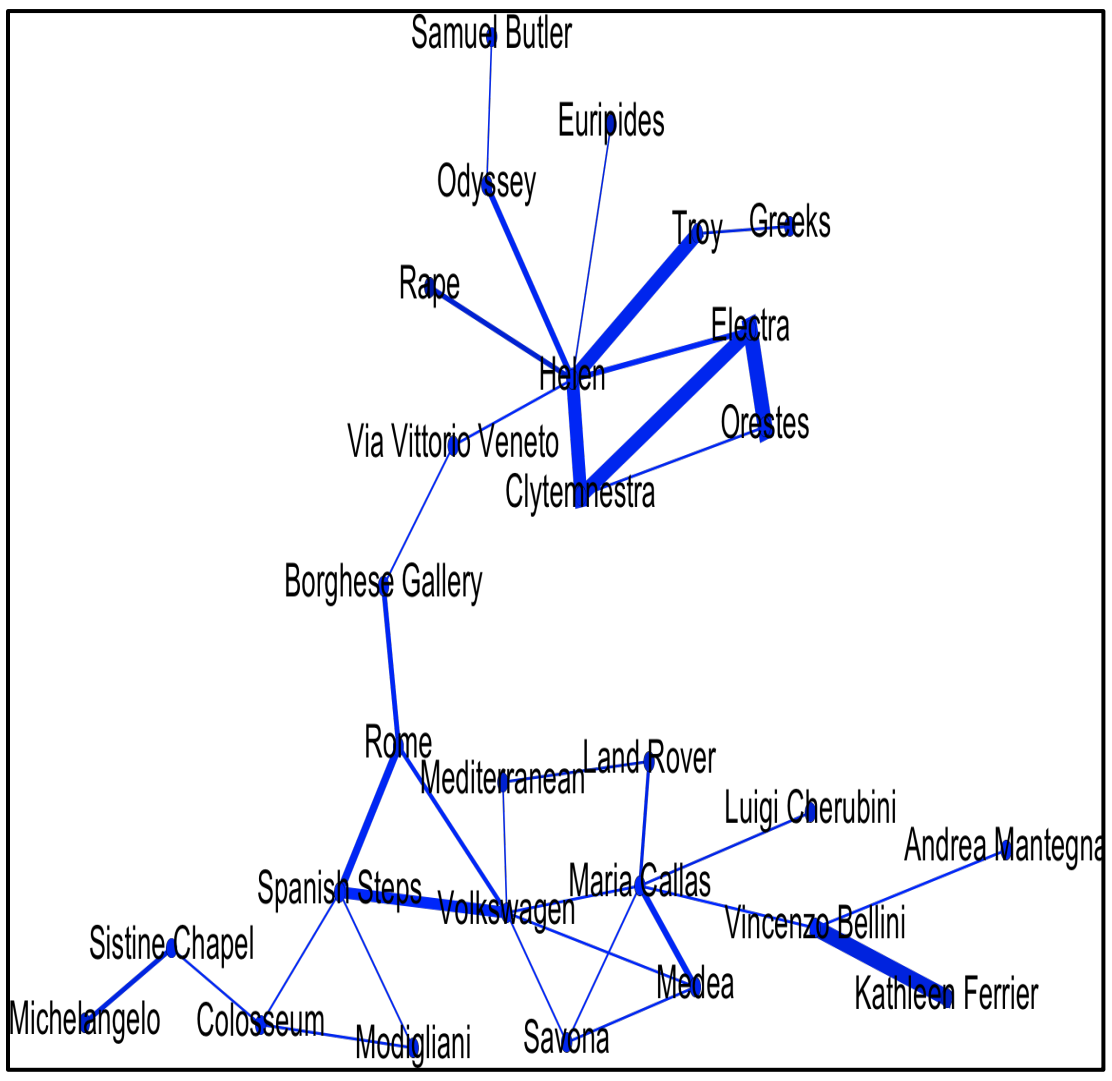

Figure 3: The Rome Network

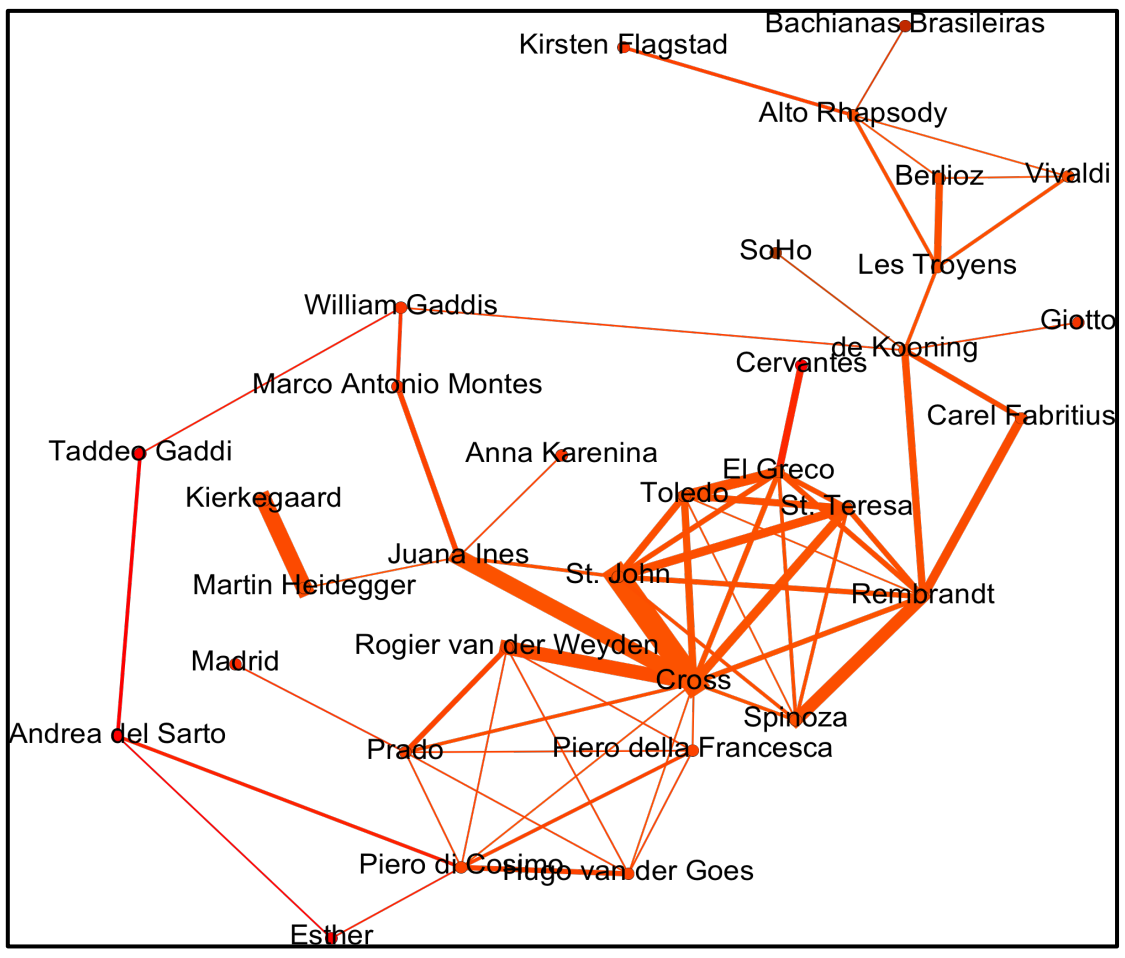

Figure 4: The Spanish Network

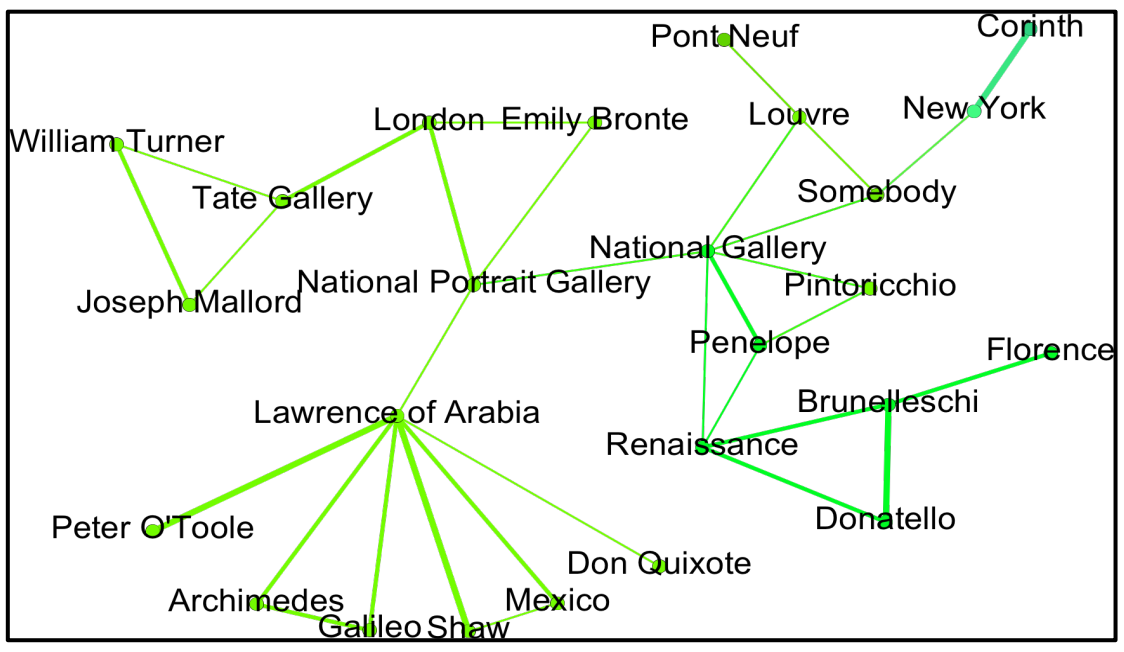

Figure 5: The Gallery Network

Wallace's (1990) contention was that the repetition in WM reflects Kate's response to a *Tractus*ized world; that she is endeavouring to connect things to establish a meaning that is absent. Our distant reading shows the obsessive persistence with which Kate pursues this attempt to connect; she appears to run over the same path of association again and again, hoping it becomes reinforced so that it persists as something "real". This analysis is consistent with one of Wallace's claims about Kate's predicament:

> *…that Kate's textual obsession is simply to find connections between things, any strands to bind the historical facts & empirical data that are all her world comprises. And always – necessarily – genuine connections elude her. All she can find is an occasional synchronicity: the fact that certain names are similar enough to be richly confusing -- William Gaddis and Taddeo Gaddi – or that certain lives & events happened to overlap in space & time. And these very thin connections turn out not to be 'real', features only of her imagination;*
>
> (Wallace, 1990, pp. 254-255)

If one examines the five disjoint networks (Figures 1-5) one notable feature of them is how they are all only *partly-thematic*, as their themes are often upset by outliers. For instance, the Homeric network (Figure 2) has mostly Homeric entities, but then Alexander the Great and Plato are thrown in (people associated by place and race but not part of the Homeric tales). Similarly, the Spanish network (Figure 4) has a lot of Spanish-related nodes (e.g., Madrid, El Greco, Prada) but then it veers into outlier sub-networks involving European philosophers and American writers. These structures confirm Wallace's proposition that Kate's networks of repetition ultimately lack real coherence and meaning, beyond that which she tries to impose by obsessive repetition. In this way, we see this network analysis as providing a highly complementary reading to that of Wallace.

## 4 The DNA of a Novel

There is one outstanding question that is invited by the current analysis, though we do not have a complete answer to it. Namely, are there definite sequences in the references made to different networks in the novel? That is, is there a tendency for a reference to one network to follow the reference to another, defining a type of higher-level DNA for the novel?

Figure 7 shows the paragraphs of the novel numbered from 1-3804 with each paragraph coloured on the basis of whether it refers to the Homeric (pink, Figure 2) or Rome Networks (blue, Figure 3). It shows a number of interesting properties: (i) these two networks dominate large parts of the novel, even though they are not the largest of the five found, (ii) there are definite sections of the novel where they dominate or fail to be referenced (i.e., the blank, white strips), (iii) references to them form definite banded sequences. To verify the latter point we analysed the frequency of these two networks in sequences of different lengths. Table 3 shows that they tend to commonly follow one another, sometimes in quite long sequences.

Overall, this shows that there are higher levels of repetitive structure that could yield other insights about repetition in WM. Indeed, it may well be that the novel has definite higher-level structure (yet to be found) that could constitute the "DNA of the novel" (see Figure 8).

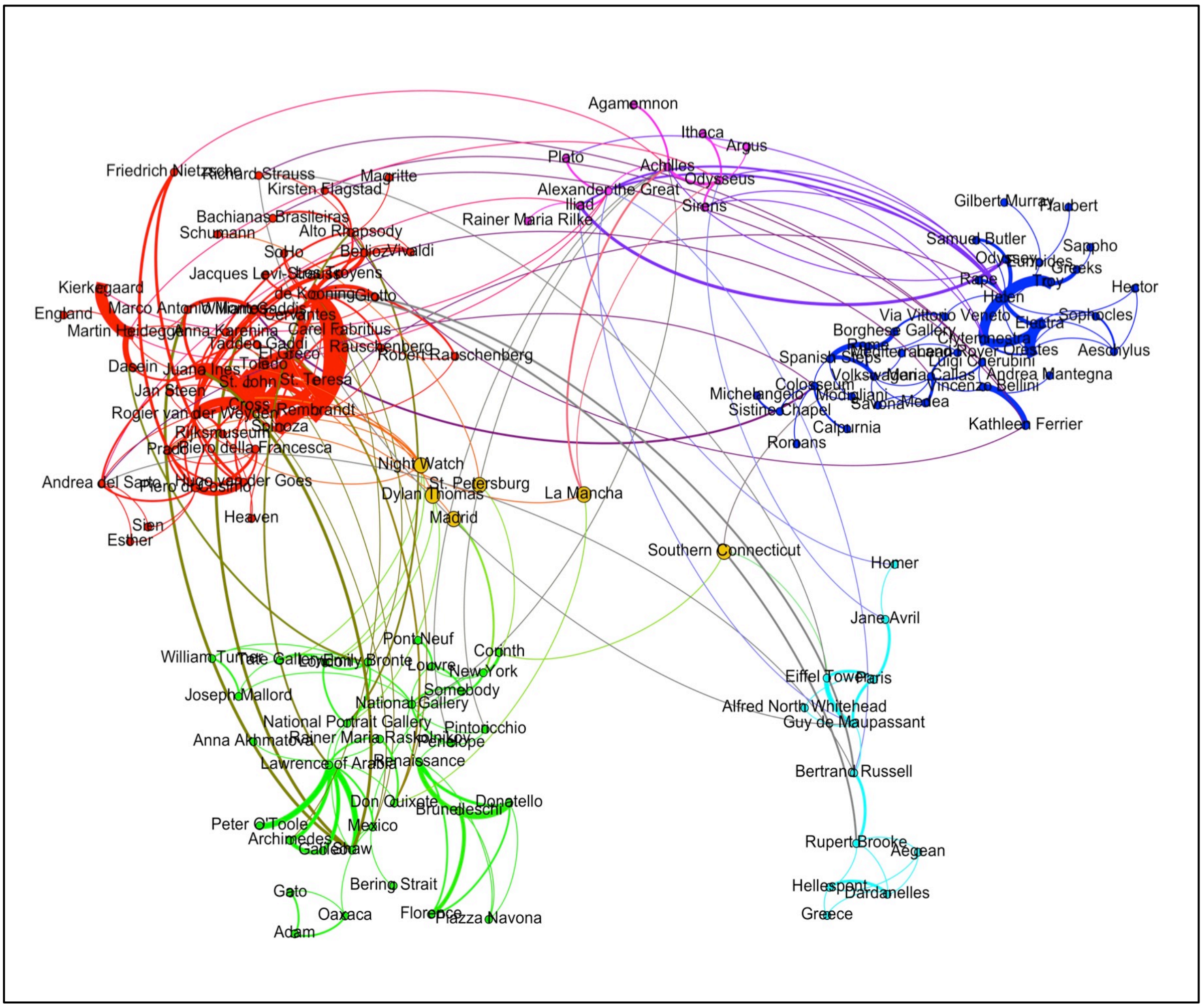

Figure 6. The Network of Networks in *Wittgenstein's Mistress*

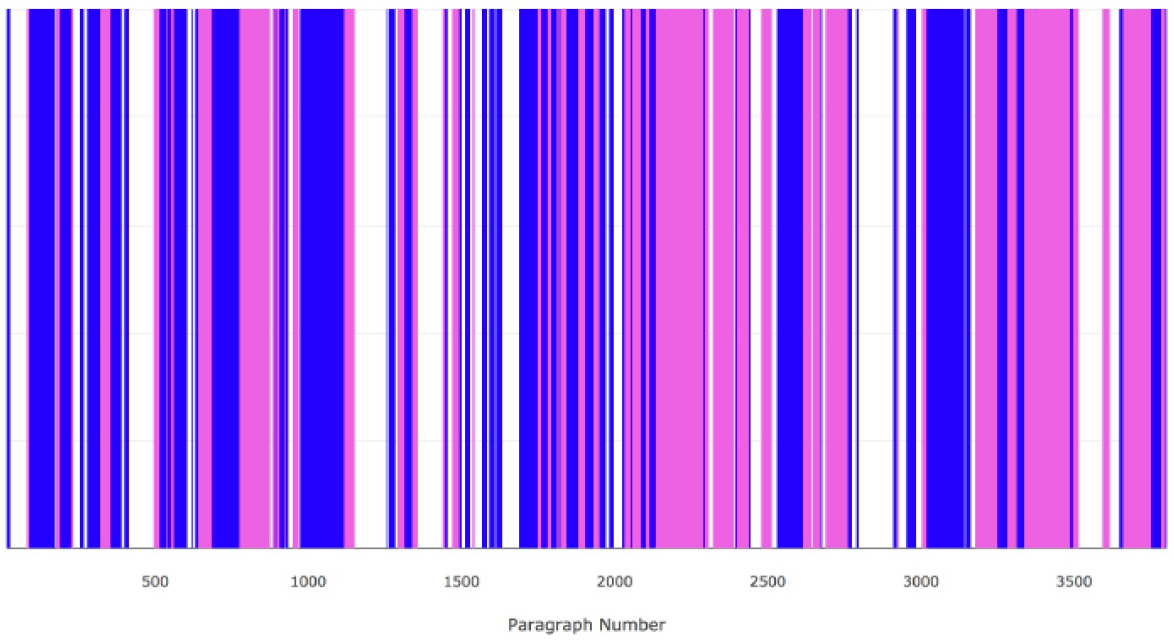

Figure 7: Paragraphs Where the Homeric (pink) and Rome Networks (blue) Are Referenced

## 5 Conclusion

In this paper, we have considered the use of data analytics techniques to provide a distant reading of David Markson's novel "Wittgenstein's Mistress". We have tried to address the role of repetition in the novel and to pursue evidence for proposals made by David Foster Wallace in his close reading of the text. In the round, specifically supporting patterns have been found in our analysis. Obviously, more could be done, in particular perhaps around the exploration of the repeating references to networks we have found; in this respect we are currently examining techniques in time-series analysis and sequence induction with this question in mind.

This analysis also invites comparative treatments of such associative references in other works that (i) are equally concise at the paragraph-level but which lack explicit repetition (e.g., Hemingway) or (ii) aim for similar stream-of-consciousness effects (e.g., Joyce). Such analyses could reveal new insights into repeated, associative referencing in diverse literary works.

## Acknowledgements

This research was an undergraduate final-year project by the first author, supervised by the second author, and was supported by Science Foundation Ireland (SFI) under Grant Number SFI/12/RC/2289.

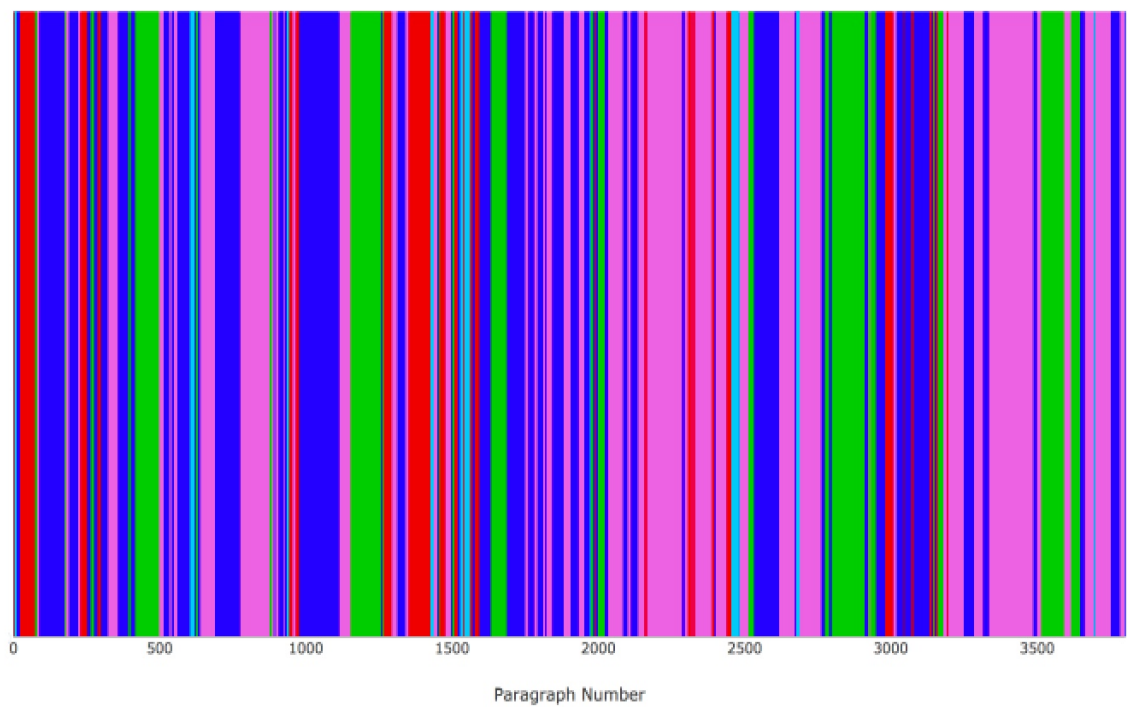

Figure 8: Paragraphs Referencing Networks: (i) Paris (cyan), (ii) Homeric (pink), (iii) Rome (blue), (iv) Spanish (red), (v) Gallery (green)

| | |
|---|---|
| <3,2> | 39 |
| <2,3> | 35 |
| <2,3,2> | 24 |
| <3,2,3> | 21 |
| <2,3,2,3> | 15 |
| <3,2,3,2> | 14 |
| <2,3,2,3,2> | 11 |
| <3,2,3,2,3> | 10 |
| <2,3,2,3,2,3> | 8 |
| <3,2,3,2,3,2> | 9 |
| <2,3,2,3,2,3,2> | 7 |
| <3,2,3,2,3,2,3> | 6 |
| <2,3,2,3,2,3,2,3> | 4 |
| <3,2,3,2,3,2,3,2> | 5 |
| <2,3,2,3,2,3,2,3,2,3> | 2 |
| <3,2,3,2,3,2,3,2,3,2> | 2 |

Table 3: Frequencies of Paragraph Sequences Referencing Homeric (2) & Rome Networks (3)